\documentclass[runningheads]{llncs}

\usepackage{graphicx}
\usepackage{booktabs}

\usepackage{booktabs}          
\usepackage{multirow}          
\usepackage[table]{xcolor}     

\definecolor{bestblue}{RGB}{180,200,235}    
\definecolor{secondblue}{RGB}{205,220,245}  
\definecolor{thirdblue}{RGB}{225,235,250}   

\newcommand{\best}[1]{\cellcolor{bestblue}\textbf{#1}}
\newcommand{\second}[1]{\cellcolor{secondblue}{#1}}
\newcommand{\third}[1]{\cellcolor{thirdblue}{#1}}

\usepackage{caption}

\usepackage{tabularx}
\usepackage{booktabs}
\usepackage{graphicx}
\usepackage{fontawesome5} 

\definecolor{medRed}{RGB}{255, 228, 225}    
\definecolor{Red}{RGB}{220, 20, 60}         
\definecolor{darkRed}{RGB}{139, 0, 0}       
\definecolor{medBlue}{RGB}{235, 242, 255}   
\definecolor{RoyalBlue}{RGB}{65, 105, 225}  
\definecolor{darkBlue}{RGB}{0, 0, 139}      
\definecolor{gtBg}{RGB}{245, 245, 245}      

\newcommand{\hlfail}[1]{%
  \begingroup\setlength\fboxsep{1pt}%
  \colorbox{medRed}{%
    \textcolor{Red}{\faTimesCircle} \textcolor{darkRed}{\textbf{#1}}%
  }\endgroup%
}

\newcommand{\hlpass}[1]{%
  \begingroup\setlength\fboxsep{1pt}%
  \colorbox{medBlue}{%
    \textcolor{RoyalBlue}{\faCheckCircle} \textcolor{darkBlue}{\textbf{#1}}%
  }\endgroup%
}

\newcommand{\hlref}[1]{%
  \begingroup\setlength\fboxsep{1pt}%
  \colorbox{medBlue}{\textcolor{darkBlue}{\textbf{#1}}}\endgroup%
}

\usepackage[table]{xcolor}
\usepackage{tabularx}
\usepackage{array}
\definecolor{tblgray}{RGB}{245,245,245} 

\usepackage{capt-of}   
\usepackage{placeins}  

\usepackage{amsmath,amssymb}

\usepackage{hyperref}
\usepackage{cleveref}

\usepackage{orcidlink}

\begin{document}

\title{DiffVP: Differential Visual Semantic Prompting for LLM-Based CT Report Generation} 

\titlerunning{DiffVP for LLM-Based CT Report Generation}





\author{
Yuhe Tian\inst{1,3,4} \and
Kun Zhang\inst{2,3,4*} \and
Haoran Ma\inst{2,3,4} \and
Rui Yan\inst{2,3,4} \and
Yingtai Li\inst{2,3,4} \and
Rongsheng Wang\inst{2,3,4} \and
Shaohua Kevin Zhou\inst{2,3,4,5*}
}

\authorrunning{Y.~Tian et al.}

\institute{
Department of Electronic Engineering and Information Science,
School of Information Science and Technology,
University of Science and Technology of China (USTC),
Hefei, Anhui 230026, China
\and
School of Biomedical Engineering, Division of Life Sciences and Medicine, University of Science and Technology of China (USTC), Hefei, Anhui 230026, China
\and
Center for Medical Imaging, Robotics, Analytic Computing \& Learning (MIRACLE),
Suzhou Institute for Advanced Research,
University of Science and Technology of China (USTC),
Suzhou, Jiangsu 215123, China
\and
Jiangsu Provincial Key Laboratory of Multimodal Digital Twin Technology,
University of Science and Technology of China (USTC),
Suzhou, Jiangsu 215123, China
\and
State Key Laboratory of Precision and Intelligent Chemistry,
University of Science and Technology of China (USTC),
Hefei, Anhui 230026, China
}

\maketitle


\begin{abstract}

While large language models (LLMs) have advanced CT report generation, existing methods typically encode 3D volumes holistically, failing to distinguish informative cues from redundant anatomical background. Inspired by radiological cognitive subtraction, we propose Differential Visual Prompting (DiffVP), which conditions report generation on explicit, high-level semantic scan-to-reference differences rather than solely on absolute visual features. DiffVP employs a hierarchical difference extractor to capture complementary global and local semantic discrepancies into a shared latent space, along with a difference-to-prompt generator that transforms these signals into learnable visual prefix tokens for LLM conditioning. These difference prompts serve as structured conditioning signals that implicitly suppress invariant anatomy while amplifying diagnostically relevant visual evidence, thereby facilitating accurate report generation without explicit lesion localization.
On two large-scale benchmarks, DiffVP consistently outperforms prior methods, improving the average BLEU-1-4 by +10.98 and +4.36, respectively, and further boosts clinical efficacy on RadGenome-ChestCT (F1 score 0.421). All codes will be released at \url{https://github.com/ArielTYH/DiffVP/}.

\end{abstract}
\section{Introduction}
\label{sec:intro}

\begin{figure}[t]
    \centering
    \includegraphics[width=\linewidth]{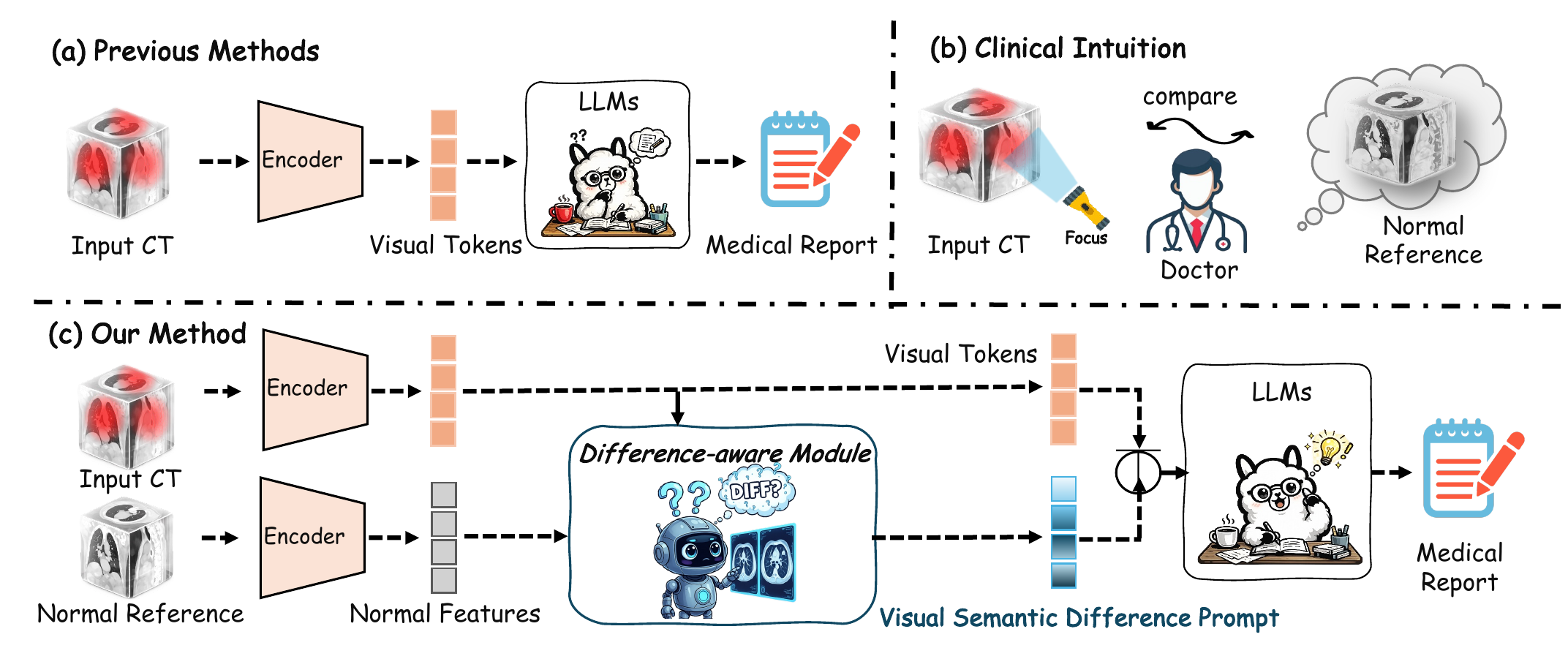}
    \caption{ Comparison of conventional methods and our proposed framework. (a) Previous methods feed whole-volume visual tokens uniformly into an LLM for report generation. (b) In clinical practice, radiologists often compare the current scan image with a normal reference image to identify differences. (c) Inspired by this, our method leverages a normal reference prior to derive deviation-aware visual prompts, guiding the LLM to generate more accurate and fine-grained reports.}
    \label{fig:teaser}
\end{figure}

Computed tomography (CT) is one of the most widely used imaging modalities in clinical practice, which plays a critical role in disease screening, diagnosis, and treatment planning across a broad range of conditions.
Despite its clinical value, interpreting CT scans remains highly demanding for radiologists, requiring careful slice-by-slice review to produce reliable reports. 
This challenge is particularly pronounced due to the inherently sparse nature of 3D CT volumes, where abnormal areas constitute only a small portion of the entire volume. 
Consequently, automated report generation, as an effective solution to assist clinicians, has attracted widespread research attention.

Early studies on CT report generation predominantly adopt an encoder-decoder paradigm~\cite{wang2021self,chen2020generating,wang2022medical,hamamci2024ct2rep}, in which a visual encoder extracts representations from CT slices or volumes, followed by a language decoder that generates reports conditioned on these visual features.
More recently, driven by the remarkable reasoning and language generation capabilities of large language models (LLMs), the field has shifted toward LLM-based medical report generation. 
R2GenGPT first introduced LLMs to this task~\cite{wang2023r2gengpt}, and Reg2RG further improves it by incorporating segmented local regions to provide finer-grained visual context~\cite{chen2025large}.
As depicted in \cref{fig:teaser}(a), existing LLM-based CT report generation typically encodes a CT volume into a sequence of visual tokens and feeds them into an LLM to generate the report.
This paradigm typically passes all tokens to the LLM uniformly, implicitly assuming equal informativeness, and relies on the LLM to autonomously infer clinically salient cues from raw visual representations.

However, this uniform token-feeding strategy is fundamentally misaligned with the actual cognitive process of radiologists when composing CT reports: in clinical practice, radiologists rely on strong priors of normal anatomy and summarize imaging findings by contrasting the current scan with an expected internal reference~\cite{kundel1975interpreting}. In fact, this comparison is not performed at the pixel level; due to inherent inter-subject anatomical variations and inevitable spatial registration errors, raw scans lack direct comparability at the pixel scale, as shown in \cref{fig:heatmap}(b).
Furthermore, this comparative process does not require explicit contour delineation or precise localization of abnormal regions; instead, it serves as an abstract, high-level semantic cue designed to highlight deviations worthy of clinical attention.

In contrast, existing LLM-based paradigms provide no such visual reference signal, leading to two key issues. First, 3D CT volumes are dominated by redundant background anatomy (e.g., bones and muscles). Without reference-guided cues, the LLM must process these undifferentiated tokens equally, which can dilute clinically important information and hinder effective report generation. Second, abnormalities in 3D CT are often sparse. Radiologists identify them by leveraging normal anatomical priors, whereas LLMs conditioned on a single scan lack this contrastive reference, making it difficult to emphasize diagnostically relevant findings while avoiding redundancy- and sparsity-induced noise. As a result, the generated reports tend to be overly generic and often miss fine-grained descriptive details that are critical for clinical interpretability.

Drawing inspiration from this semantic comparative reasoning mechanism, we propose \textbf{Diff}erential \textbf{V}isual \textbf{P}rompting for CT Report Generation (\textbf{DiffVP}).
Instead of explicitly localizing or segmenting lesions, DiffVP leverages normal CT scans as a reference prior and extracts semantic deviation-aware cues to guide attention toward clinically meaningful differences.
By contrasting the input scan with a normal reference, DiffVP constructs a visual difference prompt that reweights the conditioning semantics, implicitly suppressing invariant anatomy while amplifying distinctive variations, promoting more fine-grained and informative report generation.
Specifically, DiffVP comprises two modules.
First, the Hierarchical Difference Extractor (HDE) projects both target and reference normal inputs into a unified, fixed-budget token space via a shared extractor. Rather than relying on alignment-sensitive pixel-level subtraction, it computes high-level semantic difference representations using hierarchically complementary global and local difference operators.
This step effectively disentangles potentially diseased signals from the anatomical surroundings while resisting registration-induced noise.
Second, the Difference-to-Prompt Generator (DPG) transforms these difference representations into learnable prefix\cite{li2021prefix} embeddings.
These embeddings are then integrated into the LLM input sequence, serving as continuous cues that guide the model to generate diagnostic-sensitive reports.




Our main contributions are summarized as follows:
\begin{itemize}
    \item We propose the Differential Visual Prompting (DiffVP) framework, a novel paradigm that introduces normal CT scans as visual reference priors for CT report generation. Unlike conventional methods that rely solely on absolute visual representations or are sensitive to low-level misalignment when forming comparisons, DiffVP models differences in a high-level semantic space. This enables LLMs to better focus on clinically important information without explicit abnormal region localization.
    \item We design a hierarchical difference extractor to capture complementary global and local semantic discrepancies, together with a difference-to-prompt generator that projects these signals into the LLM's input space. This design effectively alleviates redundancy and sparsity issues inherent in 3D CT volumes while preserving diagnostic sensitivity and improving robustness to inter-subject anatomical variation.
    \item We conduct extensive experiments on two large-scale 3D CT datasets, Rad Genome-ChestCT and CTRG-Chest-548K. Results demonstrate that DiffVP consistently outperforms state-of-the-art methods across major metrics, producing reports with improved clinical accuracy, diagnostic relevance, and fine-grained clinical descriptions.
\end{itemize}

\section{Related Work}
\label{sec:related}

\subsection{3D CT Report Generation}
Early studies on 3D CT report generation, such as CT2REP and CT-ViT~\cite{hamamci2024ct2rep,marcos2024pure}, largely follow encoder--decoder architectures that compress volumetric data into global representations. Such compression inevitably discards subtle pathological cues, limiting the clinical fidelity of the generated reports.

With the strong generation and reasoning abilities of large language models (LLMs), recent work increasingly adapts LLMs for medical report generation. Generalist foundation models, including RadFM~\cite{wu2025towards}, M3D~\cite{bai2024m3d}, and Argus~\cite{liu2025argus}, use lightweight adapters to connect 3D visual encoders with pre-trained LLMs. To improve diagnostic precision, several methods further inject fine-grained guidance: Reg2RG~\cite{chen2025large} leverages anatomical region masks to extract region-aware features and align structures with text, while Dia-LLaMA~\cite{chen2025dia} introduces diagnostic prompts to steer decoding. Other lines explore more elaborate pipelines, such as the agent-based CT-Agent~\cite{mao2025ct} and the retrieval-augmented MvKeTR~\cite{deng2025mvketr}. Despite these advances, most methods still operate in a single-scan setting, extracting features only from the target volume. Lacking a normal reference for contrast, they struggle to suppress anatomical redundancy and background noise, which hinders isolating pathological changes from normal anatomy.

\subsection{Difference-Aware Report Generation}

Prior work has explored difference-aware strategies with distinct calculation and utilization paradigms. In longitudinal settings, Yun~et al.~\cite{yun2025diff} explicitly extracts isolated and localized disease-wise patch differences between current and prior scans to capture disease progression, while Song~et al.~\cite{song2025ddatr} computes dynamic temporal residuals, treating them as implicit features to emphasize evolving regions. Several methods have also addressed abnormal---normal discrepancies in non-temporal settings. For instance, Park~et al.~\cite{park2020feature} and Lyu~et al.~\cite{lyu2024automatic} perform simple, direct feature-vector subtraction between abnormal and normal 2D images. The subtracted difference vectors are then fused with tag information and fed into traditional decoders for text generation. Alternatively, rather than employing paired reference images, Zhao~et al.~\cite{zhao2023normal} constructs a latent decoupling memory matrix to learn and separate normal from pathological patterns across the entire dataset.


However, applying these mechanisms to LLM-based 3D report generation reveals numerous limitations. At the calculation level, extracting isolated patches or performing naive subtraction destroys cross-slice continuity, subsequently leading to the omission of minute lesions and introducing spatial noise. Furthermore, these differences are typically relegated to mere implicit auxiliary features. To effectively handle the inherent lesion sparsity and massive background redundancy in 3D CT volumes, our method hierarchically models discrepancies directly over the full 3D voxel-token sequence. This multi-scale approach preserves volumetric continuity and enables the fine-grained detection of structural deviations, effectively disentangling critical anatomy from the background. After extracting the hierarchical visual semantic differences, we innovatively and explicitly project them into learnable visual prefix tokens to serve as part of the LLM input prefix. This prefix-based injection provides a structured and controllable autoregressive signal, guiding the LLM to focus more attention on what differs between scans.
\section{Methodology}

\subsection{Overview}

\begin{figure}[t]
  \centering
  \includegraphics[width=\linewidth]{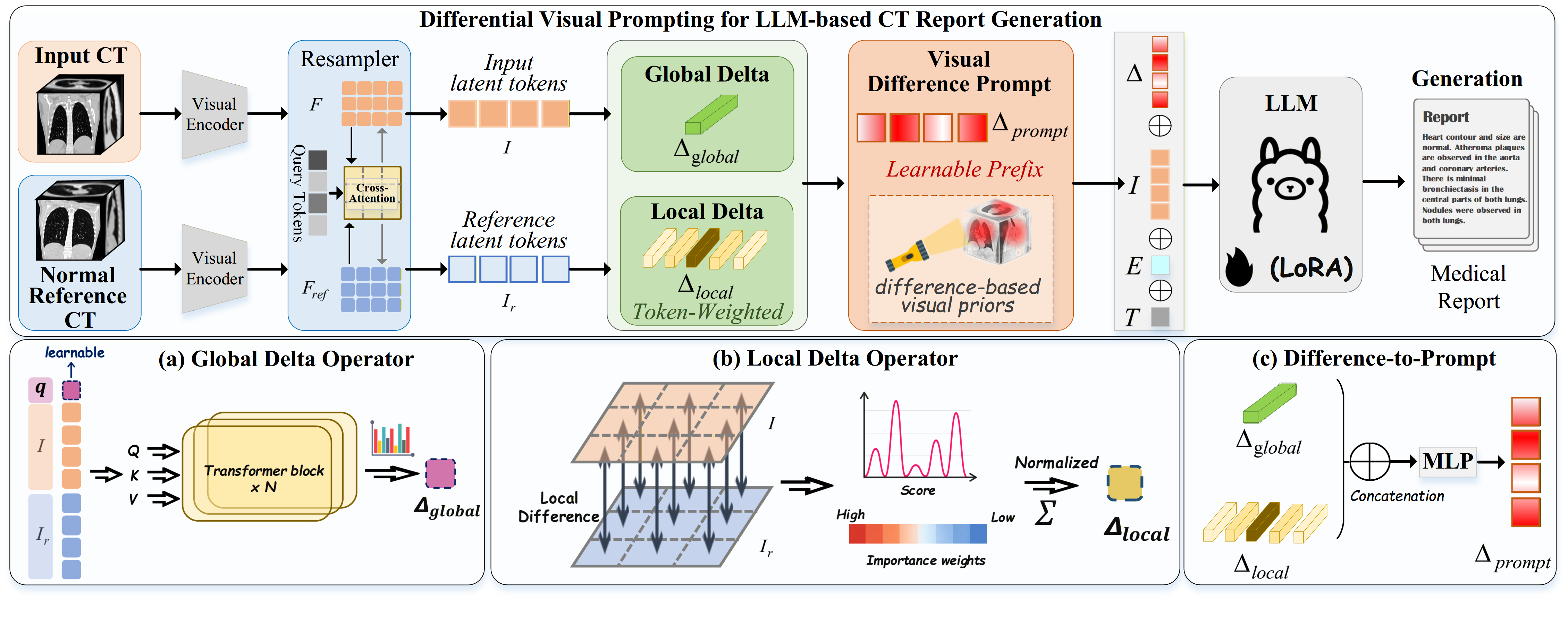}
  \caption{
  \textbf{Overall architecture.}
The framework takes a target CT volume and a normal reference CT as inputs. A shared visual encoder and resampler produce aligned latent tokens ($I$ and $I^{r}$). A difference-aware module then derives (a) a global delta $\Delta_{\text{global}}$ by applying a Transformer with a learnable query over $I$ and $I^{r}$, and (b) a local delta $\Delta_{\text{local}}$ by aggregating token-wise residuals with distance-based importance weights. These two signals are fused by (c) a difference-to-prompt generator to form a visual difference prompt, which is prepended as a soft prefix to an LLM (LoRA-tuned) to guide medical report generation.
}
  \label{fig:framework}
\end{figure}

{Existing LLM-based CT report generation} typically follows a unified paradigm: the input CT volume $X$ is first transformed into a sequence of visual tokens $I \in \{v_i\}_{i=1}^{k}$ ($k$ denotes the number of visual tokens), which are then directly fed into the LLM to generate the diagnostic report $\hat{T}$:
\begin{equation}
T \sim \text{LLM}(\hat{T} \mid I), \ I  = \Phi(X),
\label{eq:baseline_overview}
\end{equation}
where $\Phi(\cdot)$ denotes the visual encoder. While effective to some extent, this paradigm fundamentally treats all visual tokens as equally informative and relies on the LLM to implicitly discover clinically relevant cues from the input CT.

In contrast, our {Differential Visual Prompting CT report generation} (\textbf{DiffVP}) explicitly models the discrepancies between the target input $X$ and a normative reference $X_{ref}$ in a shared latent space.
The extracted difference features are then transformed into learnable prefix embeddings and prepended to the LLM input sequence as:
\begin{equation}
\label{eq:ours_overview}
\begin{aligned}
T &\sim \text{LLM}\bigl(\hat{T} \mid [\Delta_{\text{prompt}};\, I]\bigr), \\
& \text{s.t.}\quad 
 \Delta_{\text{prompt}} = \mathrm{Diff}\bigl(\Phi(X),\, \Phi(X_{\text{ref}})\bigr),
\end{aligned}
\end{equation}
where $\Delta_{\text{prompt}}$ denotes the learnable prompt tokens derived from the feature disparity between $\Phi(X)$ and $\Phi(X_{ref})$, functioning as discriminative semantic cues. 
$\mathrm{Diff}(\cdot)$ is a difference module that captures visual changes between the target and reference 3D CT volumes. 
By injecting explicit differential cues, DiffVP effectively mitigates interference from the redundant anatomical background, steering the LLM to prioritize the description of potentially important areas to improve report generation quality.
In the following, we introduce the visual encoder (\cref{sec:visual_encoding}), the hierarchical difference extractor (\cref{sec:delta_global} and \cref{sec:local_delta}), and the difference-to-prompt generator (\cref{DPG}) with LLM input construction.

\subsection{Visual Encoder}
\label{sec:visual_encoding}
To extract dense anatomical representations from high-dimensional CT volumes,
we employ a ResNet-based visual encoder as the backbone network.
Given a batch of input volumes
$X \in \mathbb{R}^{b \times s \times h \times w}$, the encoder processes the volumetric data to capture voxel-level visual patterns.
The resulting feature maps are flattened along the spatial dimensions and
denoted as
$F \in \mathbb{R}^{b \times v \times c}$,
where $v$ corresponds to the number of voxel tokens and $c$ is the encoder feature dimension.
These dense volumetric features serve as the input for subsequent difference modeling.

\subsection{Hierarchical Difference Extractor}

We aim to capture visual differences between the input and reference scans to guide the LLM in report generation. In 3D CT volumes, organ- and structure-level variations manifest at multiple spatial scales, ranging from global morphological shifts to subtle localized deviations within specific anatomical regions. Therefore, a single-scale comparison is often insufficient to represent such heterogeneous differences. To address this, we design the Hierarchical Difference Extractor (HDE) to model discrepancies at two complementary levels. Specifically, a \textit{Global Delta Operator} captures holistic structural and semantic shifts, while a \textit{Local Delta Operator} aggregates fine-grained deviations, jointly providing multi-scale difference cues for report generation.

\subsubsection{Global Delta Operator}
\label{sec:delta_global}

Our objective is to capture global differential semantics between a target CT volume $X$ and a normative reference $X_{\text{ref}}$, enabling the model to reason about \emph{what has changed} at a holistic level rather than processing each input independently. 

Both volumes are first encoded by a 3D visual encoder into dense volumetric feature maps,
$F, F_{\text{ref}} \in \mathbb{R}^{b \times v \times c}$.
However, due to the prohibitive magnitude of the voxel count $v$, directly operating on these features is computationally infeasible and introduces substantial redundancy, as most voxels correspond to common anatomical background with limited differential information.

Specifically, we adopt a Q-Former--style Visual Resampler to perform semantic downsampling. We initialize a shared set of $n$ learnable latent queries $L \in \mathbb{R}^{n \times d}$, which serve as semantic anchors for both the target and reference volumes. Using cross-attention, these shared queries aggregate informative visual content from the dense features into compact latent representations:
\begin{equation}
\begin{aligned}
I &= \text{Resampler}(F, L) \in \mathbb{R}^{b \times n \times d}, \\
I^{r} &= \text{Resampler}(F_{\text{ref}}, L) \in \mathbb{R}^{b \times n \times d},
\end{aligned}
\end{equation}
By enforcing a shared query set, this resampling process projects both volumes into a common latent space, ensuring structural alignment and enabling direct comparison at the token level.

Given the resampled latent tokens $I$ and $I^{r}$, we then explicitly model their global difference semantics. We introduce a learnable global difference query $\delta \in \mathbb{R}^{d}$, which is designed to summarize holistic discrepancies between the target and reference volumes. This query interacts jointly with both token sets through a Transformer layer:
\begin{equation}
\label{eq:delta_operator}
\begin{aligned}
\Delta_{\text{global}}
&= \mathcal{D}_{diff}\!\left(I, I^{r}\right) \\
&= \operatorname{Transformer}_{\theta}\big([\delta;\, I;\, I^{r}]\big)_{\delta},
\end{aligned}
\end{equation}
Here, $[\cdot;\cdot]$ denotes concatenation along the token dimension, and $(\cdot)_{\delta}$ indicates extracting the output state corresponding to the query token $\delta$. The resulting $\Delta_{\text{global}}$ encapsulates the overall structural and semantic shift between $X$ and $X_{\text{ref}}$, providing a compact global difference representation to condition downstream report generation.

\subsubsection{Local Delta Operator}
\label{sec:local_delta}

While $\Delta_{\text{global}}$ captures high-level semantic discrepancies, it may not explicitly encode the strength of localized deviations. To complement it, we introduce a \emph{local-level difference aggregation} mechanism that operates directly on the aligned latent visual tokens.

Formally, given the input and reference tokens $I, I^{r}$, we define a local difference operator $\mathcal{A}_{\text{diff}}$ as
\begin{equation}
\Delta_{\text{local}}
=
\mathcal{A}_{\text{diff}}\!\left(I, I^{r}\right),
\label{eq:local_delta_operator}
\end{equation}
where $\Delta_{\text{local}} \in \mathbb{R}^{b \times d}$ summarizes localized deviations via a geometry-aware weighting scheme in the latent space.

Specifically, we assign each aligned token pair $(I_i, I_i^{r})$ a normalized importance weight $w_i$ based on their squared Euclidean distance:
\begin{equation}
w_i
=
\frac{\left\lVert I_i - I^{r}_i \right\rVert_2^2}
{\sum_{j=1}^{N} \left\lVert I_j - I^{r}_j \right\rVert_2^2 + \epsilon},
\quad i = 1,\dots,N .
\label{eq:local_diff_weight}
\end{equation}
This weighting emphasizes token pairs that deviate most from the reference while remaining robust to scale via normalization, where $\epsilon = 1 \times 10^{-5}$ is a small term added to prevent division by zero.

The local difference representation is then obtained by aggregating the residual vectors under these weights:
\begin{equation}
\mathcal{A}_{\text{diff}}\!\left(I, I^{r}\right)
=
\sum_{i=1}^{N}
w_i \left( I_i - I^{r}_i \right).
\label{eq:local_diff_aggregation}
\end{equation}

This operator is parameter-free and grounded in the metric structure of the latent token space. It ensures that tokens exhibiting pronounced deviations contribute dominantly to $\Delta_{\text{local}}$, providing a direct and interpretable difference evidence signal for the LLM during report generation.

\subsection{Difference-to-Prompt Generator}
\label{DPG}

The HDE module yields two complementary difference representations: a global descriptor $\Delta_{\text{global}}$ and a locality-sensitive descriptor $\Delta_{\text{local}}$. The Difference-to-Prompt Generator (DPG) transforms these visual differences into a learnable prefix that can directly condition the LLM.

Since the extracted differences reside in the vision latent space, they must be projected into the LLM embedding space. To jointly encode holistic shifts and localized deviations, we first fuse $\Delta_{\text{global}}$ and $\Delta_{\text{local}}$ via concatenation and an MLP, and then expand the fused representation into a sequence of prefix tokens:
\begin{equation}
\Delta_{\text{prompt}}
=
\text{Projector}\big([\Delta_{\text{global}} \,;\, \Delta_{\text{local}}]\big).
\end{equation}

The resulting $\Delta_{\text{prompt}} \in \mathbb{R}^{p \times d_{\text{LLM}}}$ consists of $p$ learnable prefix tokens ($p{=}16$ in our study), with dimensionality matching the LLM embedding size. By injecting explicit scan-to-reference discrepancy cues initially, $\Delta_{\text{prompt}}$ acts as a dynamic soft prompt that reweights conditioning signals, suppressing background-dominated representations and enhancing report generation.

\noindent\textbf{Diagnostic Semantic Guidance.}
To provide high-level semantic guidance, we incorporate predicted diagnostic classes $E$, a standard anchoring paradigm in medical report generation~\cite{kale2023replace}. Specifically, $E$ is generated by a pre-trained 3D ResNet-18 classifier optimized via Binary Cross-Entropy (BCE) loss over 18 report-independent disease labels (e.g., Cardiomegaly, Emphysema), preventing information leakage. During generation, this classifier remains frozen, and its predictions are converted into short textual prompts. Detailed class definitions, training specifics, and the rationale for this two-stage design are deferred to the Supplementary Material.

\noindent\textbf{LLM Input Construction.}
To preserve complete visual context, we retain the original tokens $I$ with the difference prefix $\Delta_{\text{prompt}}$. We then integrate diagnostic guidance $E$ and append structured textual prompts $T$ to aid instruction following. The final input sequence for the LLM is formulated as:

\begin{equation}
\mathcal{X}_{\text{in}}
=
[\, \Delta_{\text{prompt}} \,;\, I \,;\, E \,;\, T \,].
\end{equation}

\subsection{Training Objective}
We train the model with a multi-task objective that combines report generation with an auxiliary disease classification task. 

\noindent\textbf{Report Generation Loss.}
We formulate report generation as conditional sequence modeling. Given the hybrid input embeddings $\mathcal{X}_{\text{in}}$ and the ground-truth report $Y=\{y_1,\dots,y_L\}$, we minimize the negative log-likelihood (NLL):
\begin{equation}
\mathcal{L}_{\text{gen}}
=
-\sum_{t=1}^{L} \log P\!\left(y_t \mid y_{<t}, \mathcal{X}_{\text{in}};\theta\right).
\end{equation}

\noindent\textbf{Diagnostic Classification Loss.}
To encourage diagnostic consistency and regularize the visual representations, we add an auxiliary multi-label classification loss $\mathcal{L}_{\text{cls}}$ using binary cross-entropy (BCE) over $K{=}18$ disease categories.

\noindent\textbf{Total Objective.}
The final training objective is:
\begin{equation}
\mathcal{L}_{\text{total}}
=
\mathcal{L}_{\text{gen}} + \mathcal{L}_{\text{cls}}.
\end{equation}

\section{Experiments}

\begin{table}[t]
  \centering
  \small
  \caption{Quantitative results on RadGenome-ChestCT and CTRG-Chest-548K. Models marked with $\dagger$ are foundation models.
  We report BLEU (BL), ROUGE-L (RG-L), and METEOR (MTR) scores.
  The best, second best, and third best results are highlighted in dark blue, medium blue, and light blue, respectively.}
  \label{tab:main_results}

  \resizebox{\textwidth}{!}{%
  \begin{tabular*}{\textwidth}{@{\extracolsep{\fill}}l l c cccccc}
    \toprule
    \textbf{Dataset} & \textbf{Method} & \textbf{Year} &
    \textbf{BL-1} & \textbf{BL-2} & \textbf{BL-3} & \textbf{BL-4} &
    \textbf{RG-L} & \textbf{MTR} \\
    \midrule

    \multirow{7}{*}{\shortstack{\textbf{RadGenome-}\\ \textbf{ChestCT}}}
      & $\dagger$ RadFM & 2023 & 44.20 & 34.49 & 28.06 & 23.65 & 31.53 & 39.94 \\
      & $\dagger$ M3D   & 2023 & 43.57 & 34.48 & 28.54 & 24.49 & \second{32.61} & 39.95 \\
      \cmidrule(l){2-9}
      & R2GenGPT    & 2023 & 43.28 & 34.11 & 28.16 & 24.16 & 32.26 & 39.85 \\
      & MedVInT     & 2024 & 44.28 & \third{34.91} & \third{28.75} & \third{24.60} & \third{32.58} & \third{40.39} \\
      & CT2Rep      & 2024 & \third{44.42} & 34.43 & 27.94 & 23.56 & 30.99 & 40.16 \\
      & Reg2RG      & 2025 & \second{47.25} & \second{36.49} & \second{29.57} & \second{24.87} & \best{36.65} & \second{44.07} \\
      \cmidrule(l){2-9}
      & \textbf{Ours} & -  & \best{58.16} & \best{48.74} & \best{40.93} & \best{34.28} & 32.32 & \best{47.40} \\
    \midrule

    \multirow{7}{*}{\shortstack{\textbf{CTRG-Chest-}\\ \textbf{548K}}}
      & $\dagger$ RadFM & 2023 & \third{48.66} & \third{40.28} & \third{34.73} & \third{30.89} & 49.08 & 49.18 \\
      & $\dagger$ M3D   & 2023 & 46.27 & 39.02 & 34.23 & 30.86 & \second{50.24} & 49.26 \\
      \cmidrule(l){2-9}
      & R2GenGPT    & 2023 & 41.82 & 36.37 & 32.70 & 30.10 & \best{50.93} & 47.05 \\
      & MedVInT     & 2024 & 47.38 & 39.60 & 34.28 & 30.68 & \third{49.53} & \third{49.32} \\
      & CT2Rep      & 2024 & 42.28 & 36.16 & 32.08 & 29.19 & 50.17 & 47.00 \\
      & Reg2RG      & 2025 & \second{49.63} & \second{41.43} & \second{35.91} & \second{32.04} & 47.76 & \second{49.71} \\
      \cmidrule(l){2-9}
      & \textbf{Ours} & -  & \best{50.37} & \best{46.57} & \best{41.94} & \best{37.57} & 45.00 & \best{50.44} \\
    \bottomrule
  \end{tabular*}%
  }
\end{table}

\subsection{Datasets}

\noindent\textbf{RadGenome-ChestCT.} 
The RadGenome-ChestCT dataset~\cite{zhang2024radgenome}, derived from the CT-RATE dataset~\cite{hamamci2024foundation}, comprises 25,692 CT-report pairs from 21,304 unique patients. We adhere to the official partition, designating 24,128 pairs for training and 1,564 pairs for evaluation. The data is pre-processed into ten anatomical regions with a standardized voxel spacing of $1{\times}1{\times}3$~mm.

\noindent\textbf{CTRG-Chest-548K.}
As a supplementary benchmark~\cite{tang2024work}, it contains 1,804 CT report pairs, split into train and test sets with an 8:2 ratio. We further extract anatomical region masks and parse reports into region-level descriptions.

\noindent\textbf{Normal Reference Set Construction.}
To facilitate explicit difference modeling, we construct a dedicated normal reference pool by selecting CT volumes where the associated reports indicate no abnormalities. Based on this criterion, we select 1,787 training and 124 evaluation samples from RadGenome-ChestCT, each used in its training and testing phase. Similarly, for the CTRG-Chest-548K dataset, we select 105 training and 43 testing samples as normal references.
 References are sampled strictly within the corresponding split, train-only for training, test-only for testing, preventing any train--test leakage. In the supplementary material, we further report a five-seed stability analysis with mean$\pm$std and two-sided $t$-tests under random reference pairing.

\subsection{Implementation Details}

We utilize a pretrained frozen 3D ResNet-18 as the visual encoder to extract voxel-level features ($d=512$).  For the language backbone, we adopt LLaMA-2-7B~\cite{touvron2023llama} equipped with Low-Rank Adaptation (LoRA)~\cite{hu2022lora} for efficient fine-tuning. 
All models are optimized with a learning rate of $5 \times 10^{-5}$. We train the models for 10 epochs with a batch size of 1 per GPU. Experiments are conducted on a cluster of 4 NVIDIA H20 GPUs trained in parallel.

\subsection{Evaluation Metrics}
We evaluate linguistic quality using standard generation metrics, specifically BLEU-$n$ with $n$ ranging from 1 to 4, METEOR, and ROUGE-1, ROUGE-2, and ROUGE-L, and medical correctness using Clinical Efficacy (CE) metrics. CE computes Precision, Recall, and F1 on 18 pathologies extracted by a RadBERT classifier. Due to label misalignment between RadBERT's training data and the primary abnormalities in CTRG-Chest-548K, CE metrics are reported only for RadGenome-ChestCT.


    
    
    
    
    

\begin{table}[t]
  \centering

  \begin{minipage}[t]{0.49\textwidth}
    \centering
    \small
    \captionof{table}{
      \textbf{Clinical Efficacy (CE) performance on RadGenome-ChestCT.}
      The symbol $^\dagger$ denotes generalist foundation models, while unmarked methods are specialized Report Generation Models.
    }
    \label{tab:radgenome_ce}

    \resizebox{\linewidth}{!}{%
    \begin{tabular}{l ccc}
      \toprule
      \multirow{2}{*}{\textbf{Method}} & \multicolumn{3}{c}{\textbf{Clinical Efficacy (CE)}} \\
      \cmidrule(lr){2-4}
       & \textbf{F1-Score} & \textbf{Recall} & \textbf{Precision} \\
      \midrule
      $^\dagger$RadFM     & 0.195 & 0.131 & 0.382 \\
      $^\dagger$M3D       & 0.148 & 0.090 & 0.407 \\
      \midrule
      R2GenGPT & 0.110 & 0.066 & 0.340 \\
      MedVInT  & 0.212 & 0.148 & 0.377 \\
      CT2Rep   & 0.139 & 0.089 & 0.317 \\
      Reg2RG   & 0.253 & 0.181 & \textbf{0.423} \\
      \midrule
      \textbf{Ours} & \textbf{0.421} & \textbf{0.496} & 0.366 \\
      \bottomrule
    \end{tabular}%
    }
  \end{minipage}
  \hfill
  \begin{minipage}[t]{0.49\textwidth}
    \centering
    \small
    \captionof{table}{\textbf{Impact of Visual Prefix Length.} We vary the visual prefix length $p\in\{4,8,16,24,32\}$ on RadGenome-ChestCT. NLG performance increases with $p$ and peaks at $p{=}16$ .}
    \label{tab:ablation_prefix}

    \resizebox{\linewidth}{!}{%
    \begin{tabular}{l ccccc}
      \toprule
      \multirow{2}{*}{\textbf{Metric}} & \multicolumn{5}{c}{\textbf{Prefix Length ($p$)}} \\
      \cmidrule(lr){2-6}
       & \textbf{4} & \textbf{8} & \textbf{16} & \textbf{24} & \textbf{32} \\
      \midrule
      BLEU-1  & 53.97 & 57.98 & \textbf{58.16} & 56.89 & 56.24 \\
      BLEU-2  & 44.29 & 48.60 & \textbf{48.74} & 47.46 & 46.51 \\
      BLEU-3  & 36.37 & 40.78 & \textbf{40.93} & 39.79 & 38.43 \\
      BLEU-4  & 29.72 & 34.13 & \textbf{34.28} & 33.32 & 31.53 \\
      METEOR  & 43.37 & \textbf{47.96} & 47.40 & 47.33 & 46.19 \\
      ROUGE-1 & 59.63 & 63.48 & \textbf{63.52} & 62.33 & 61.82 \\
      ROUGE-2 & 38.41 & 42.60 & \textbf{42.63} & 41.49 & 40.38 \\
      ROUGE-L & 31.46 & \textbf{32.43} & 32.32 & 32.27 & 31.23 \\
      \bottomrule
    \end{tabular}%
    }
  \end{minipage}

\end{table}

\subsection{Comparison with SOTA Methods}
\paragraph{Natural Language Generation Performance.}
Table~\ref{tab:main_results} summarizes the NLG results on RadGenome-ChestCT and CTRG-Chest-548K. We compare DiffVP with large-scale medical foundation models (RadFM, M3D) and dedicated report generation methods (R2GenGPT, MedVInT, CT2Rep, Reg2RG). Overall, DiffVP achieves the best or competitive performance across most metrics on both datasets. Compared with the previous SOTA Reg2RG, DiffVP delivers large gains in n-gram precision, improving BLEU-1 by 23.08\% and BLEU-4 by 37.83\%. DiffVP also attains the highest METEOR, indicating better semantic alignment with reference reports. Although Reg2RG is slightly higher on ROUGE-L, DiffVP’s stronger BLEU and METEOR suggest improved precision in describing clinically relevant findings rather than relying on long shared subsequences.

\paragraph{Clinical Efficacy Analysis.}
\Cref{tab:radgenome_ce} reports CE results on RadGenome-ChestCT. DiffVP achieves the best F1-score (0.421) and the highest Recall among all methods, indicating improved sensitivity to clinically relevant findings and a reduced risk of missing positives. While Reg2RG is marginally higher in Precision, DiffVP provides a better overall balance between Precision and Recall.

\section{Ablation Studies and Analysis}

\subsection{Ablation Studies}

\begin{table*}[t]
  \centering
  \caption{
    \label{tab:ablation_modules}
    Ablation study on the effectiveness of different modules on RadGenome-ChestCT. 
    \textbf{Global}, \textbf{Local}, and \textbf{E} denote the Global Delta, Local Delta, and Classification Prompt modules, respectively.
    CE Metrics denote clinical efficacy metrics including Precision, Recall, and F1.
  }
  \setlength{\tabcolsep}{4pt} 
  \resizebox{\textwidth}{!}{
  \begin{tabular}{l ccc cccccccc ccc} 
    \toprule
    \multirow{2}{*}{\textbf{ID}} & \multicolumn{3}{c}{\textbf{Module Settings}} & \multicolumn{8}{c}{\textbf{NLG Metrics}} & \multicolumn{3}{c}{\textbf{CE Metrics}} \\
    \cmidrule(lr){2-4} \cmidrule(lr){5-12} \cmidrule(lr){13-15}
    
     & $\Delta_{\text{global}}$  &  $\Delta_{\text{local}}$ & E & B-1 & B-2 & B-3 & B-4 & M & R-1 & R-2 & R-L & Pre. & Rec. & F1 \\
    \midrule
    
    \textbf{Baseline}
     &  &  &  & 55.29 & 46.55 & 39.36 & 33.17 & 44.11 & 62.12 & 41.95 & 31.94 & 0.295 & 0.109 & 0.159 \\
    
    \textbf{\#1}
     &  &  & \checkmark & 55.62 & 46.89 & 39.65 & 33.44 & 44.80 & 62.31 & 42.19 & 32.35 & 0.311 & 0.217 & 0.256 \\

    \textbf{\#2}
     & \checkmark &  & \checkmark & 57.46 & 47.92 & 39.97 & 33.20 & 46.63 & 62.97 & 41.69 & 32.08 & \textbf{0.373} & 0.459 & 0.411 \\

    \textbf{\#3}
     &  & \checkmark & \checkmark & 57.68 & 48.24 & 40.47 & 33.89 & \textbf{47.67} & 63.19 & 42.25 & \textbf{32.76} & 0.360 & 0.464 & 0.406 \\
    
    \midrule
    
    \textbf{Ours}
     & \checkmark & \checkmark & \checkmark & \textbf{58.16} & \textbf{48.74} & \textbf{40.93} & \textbf{34.28} & 47.40 & \textbf{63.52} & \textbf{42.63} & 32.32 & 0.366 & \textbf{0.496} & \textbf{0.421} \\
    \bottomrule
  \end{tabular}
  }

\end{table*}



\noindent\textbf{Ablation Study on Model Components.}
Table~\ref{tab:ablation_modules} reports an ablation analysis of DiffVP. We start from a baseline with a pretrained ResNet18 encoder on RadGenome-ChestCT. Adding the inferred diagnostic anchor (\#1) provides consistent improvements, notably in CE Recall, and offers a stable semantic context for subsequent difference prompting. Comparing \#2 and \#3 shows that modeling discrepancies at complementary levels is beneficial: the \textbf{Global Delta} captures holistic structural and semantic shifts and improves global consistency, while the \textbf{Local Delta} contributes fine-grained deviations that boost METEOR. Combining both yields the best BLEU-4, indicating that joint global consistency and local detail improves long-range coherence and overall report quality.

\noindent\textbf{Impact of Visual Prefix Length.}
Table~\ref{tab:ablation_prefix} studies the visual prefix length $p$. Performance increases as $p$ grows from 4, with BLEU-1--4 peaking at $p{=}16$. While $p{=}8$ remains competitive for Recall, larger prefixes (24/32) consistently degrade performance, suggesting redundant conditioning. We therefore set $p{=}16$ by default.

\begin{figure}[t]
  \centering
  \begin{minipage}[t]{0.48\textwidth}
    \centering
    \includegraphics[width=\linewidth]{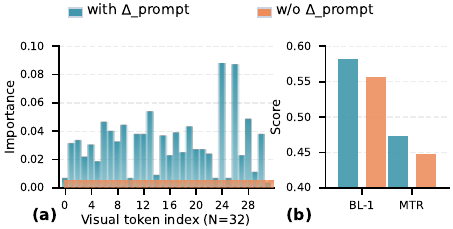}
    \caption{\textbf{Visual token importance.} (a) Normalized importance distributions for $N{=}32$ latent tokens with (blue) and without (orange) the proposed $\Delta_{\text{prompt}}$.  (b) Performance gain by $\Delta_{\text{prompt}}$.}
    \label{fig:ana}
  \end{minipage}
  \hfill 
  \begin{minipage}[t]{0.48\textwidth}
    \centering
    \includegraphics[width=\linewidth]{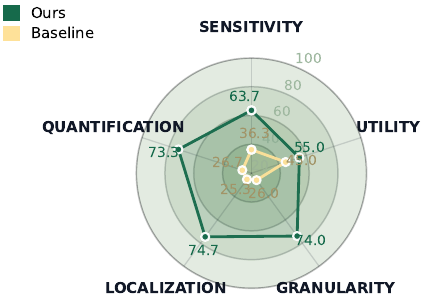}
    \caption{\textbf{LLM-as-a-Judge Evaluation}.We employ GPT-5 for pairwise comparisons across five clinical axes: Granularity, Localization, Quantification, Sensitivity, and Utility}
    \label{fig:llm}
  \end{minipage}
\end{figure}

\subsection{Performance Analysis}



\begin{figure}[t]
    \centering
    \includegraphics[width=\linewidth]{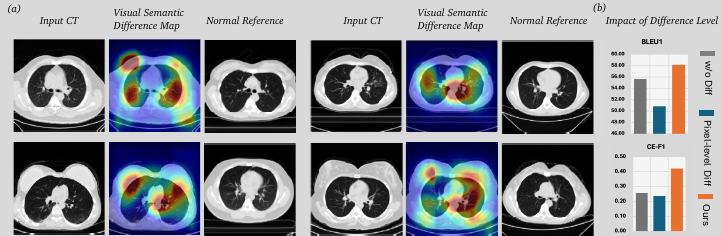} 
    \caption{\textbf{Visual semantic discrepancy and discrepancy-level comparison.}
    (a) Two examples showing the input CT, the \emph{Visual Semantic Difference Map} produced by our method, and the paired normal reference.
    (b) Performance comparison among \textit{w/o Diff}, \textit{Pixel-level Diff} (pixel-wise subtraction), and \textit{Ours} on BLEU-1 and CE-F1.}
    \label{fig:heatmap}
\end{figure}

\noindent\textbf{Visual semantic difference map.}
Fig.~\ref{fig:heatmap}(a) explicitly visualizes the critical discrepancy cue extracted by our difference-aware module.
Given an input CT scan and its paired normal reference, we compute multi-scale discrepancy in the latent visual token space and project it back to the image plane for interpretability.
This map is not obtained by simple pixel-wise subtraction; instead, it highlights semantically deviant regions relative to the normal prior, which is consistent with radiologists' high-level comparative reasoning (e.g., for the first pair in Fig.~\ref{fig:heatmap}(a), the activation concentrates around the lung fields where the report describes subsegmental atelectasis in the right lower lobe and multiple pulmonary nodules, most prominent in the left upper-lobe lingular segment).

\noindent\textbf{Impact of discrepancy level.}
Fig.~\ref{fig:heatmap}(b) compares three variants:\textit{w/o Diff}, which removes the discrepancy cue; \textit{Pixel-level Diff}, which derives the cue from pixel-wise subtraction in the image space; and \textit{Ours}, which computes discrepancy in the latent token space.
Importantly, Pixel-level Diff and Ours share the same discrepancy-prompt injection form with the same prefix structure and length, and thus differ only in where the discrepancy is computed. 
Semantic discrepancy consistently yields better BLEU-1 and CE-F1, suggesting that clinically useful comparison is better captured at a semantic level than by raw pixel subtraction.

\noindent\textbf{Analysis of the $\Delta_{\text{prompt}}$ Distribution.}
To clarify the visual prefix mechanism, \cref{fig:ana}(a) visualizes the token importance distribution, defined as the latent geometric discrepancy: $S_i = \| I_i - I_i^r \|_2^2$. This metric quantifies local clinical deviations from the reference in semantic space. 
As shown, the baseline (orange) exhibits a near-uniform distribution, failing to distinguish critical findings from massive anatomical redundancy. In sharp contrast, $\Delta_{\text{prompt}}$ (blue) induces a highly sparse distribution, where the top 8 tokens capture 45\% of the total importance mass. This shift proves our prefix forces the model to focus on localized clinically-variant regions rather than treating all tokens equally. Such difference-guided focus aligns with the linguistic gains in \cref{fig:ana}(b), underscoring the necessity of prioritizing semantically variant regions for high-quality report generation.

\begin{figure}[t]
    \centering
    \scriptsize 
    \setlength{\tabcolsep}{0pt} 
    \renewcommand{\arraystretch}{1.1} 

    \renewcommand{\tabularxcolumn}[1]{m{#1}}

    \begin{tabularx}{\linewidth}{ 
        @{} 
        m{0.13\linewidth}        
        @{\hspace{4pt}}          
        >{\centering\arraybackslash}m{2.5em} 
        @{\hspace{4pt}}          
        >{\raggedright\arraybackslash}X      
        @{} 
    }
        \toprule
        \multicolumn{1}{c}{\textbf{Visuals}} & & \multicolumn{1}{l}{\textbf{Excerpt of the generated report}} \\
        \midrule

        \includegraphics[width=\linewidth, keepaspectratio]{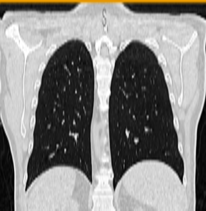} 
        & \cellcolor{gray!15}\rotatebox{90}{\textbf{GT}} 
        & \cellcolor{gtBg}%
          \textbf{Abdomen:} \hlref{A decrease in liver density, consistent with steatosis}, is observed. Bilateral adrenal glands were normal and no space-occupying lesion was detected. Thoracic aorta diameter is normal. \par
          \textbf{Lung:} Several \hlref{nonspecific nodules} are observed in both lungs, the largest of which is in the left lung lower lobe anteromedial segment (\hlref{4.6 mm}). There was no finding in favor of active infiltration in both lungs. \\
        \midrule

        \includegraphics[width=\linewidth, keepaspectratio]{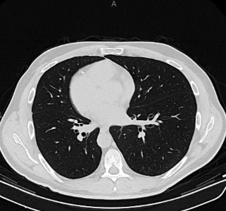} 
        & \cellcolor{medRed}\rotatebox{90}{%
            \begin{tabular}{@{}c@{}}
                \textbf{\textcolor{darkRed}{Comparison}} \\
                \textbf{\textcolor{darkRed}{Method}}
            \end{tabular}%
          } 
        & \textbf{Abdomen:} Bilateral adrenal glands were normal and \hlfail{no space-occupying lesion was detected}. Upper abdominal organs included in the sections are \hlfail{normal}. \textit{[Missed Steatosis]} \par
          \textbf{Lung:} Trachea and both main bronchi are open. The lungs are \hlfail{clear of active infiltration}. \textit{[Missed Nodules]} \\
        \midrule

        \includegraphics[width=\linewidth, keepaspectratio]{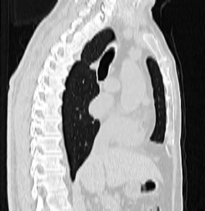} 
        & \cellcolor{medBlue}\rotatebox{90}{%
            \begin{tabular}{@{}c@{}}
                \textbf{\textcolor{darkBlue}{Ours}} \\
                \textbf{\textcolor{darkBlue}{(w/o $\Delta$ prompt)}}
            \end{tabular}%
          } 
        & \cellcolor{medBlue!10}%
          \textbf{Abdomen:} Bilateral adrenal glands were normal. Upper abdominal organs included in the sections are \hlfail{normal}. \textit{[Missed Steatosis]} No space-occupying lesion was detected in the liver. \par
          \textbf{Lung:} \hlfail{No mass lesion} with distinguishable borders was detected. \textit{[Missed Nodule]} However, when examined in the parenchyma window, \hlpass{Linear subsegmental atelectasis was observed} in the middle lobe of the right lung. \textit{[Fine-grained detail]} \\
        \midrule

        \includegraphics[width=\linewidth, keepaspectratio]{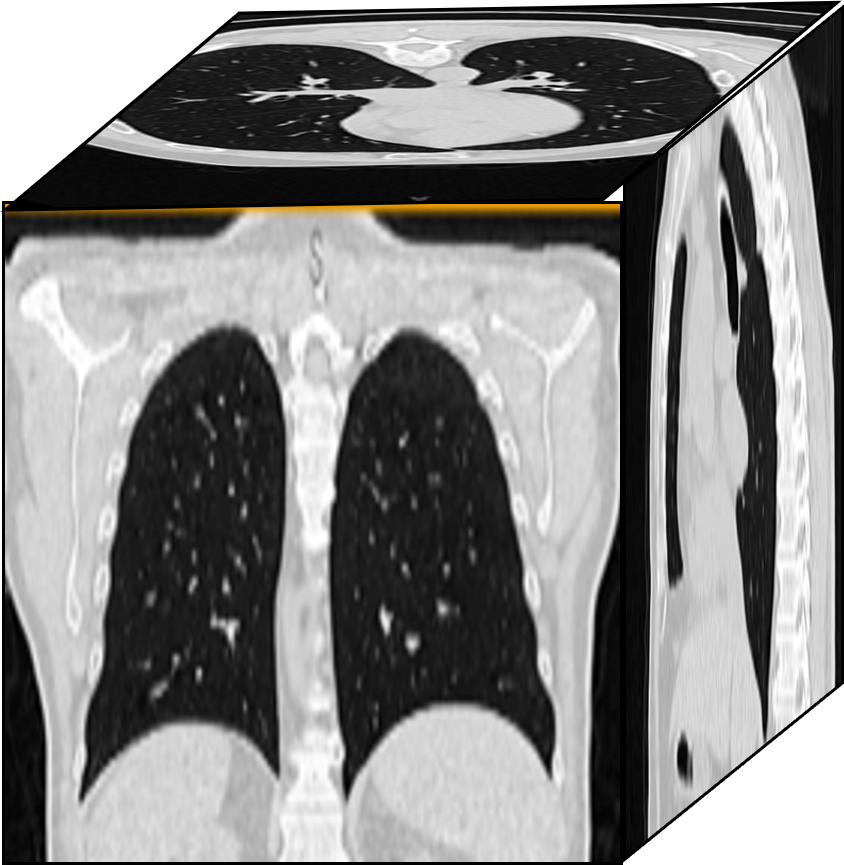} 
        & \cellcolor{medBlue}\rotatebox{90}{\textbf{\textcolor{darkBlue}{Ours}}} 
        & \cellcolor{medBlue!10}%
          \textbf{Abdomen:} A \hlpass{decrease in liver density} is noted, consistent with \hlpass{steatosis}. Bilateral adrenal glands normal and no space-occupying lesion was detected. \par
          \textbf{Lung:} \hlpass{Nonspecific nodules} are seen in the lung parenchyma. The largest nodule is measured at \hlpass{4.6 mm} in the left lower lobe. There was no finding in favor of active infiltration. \\
        
        \bottomrule
    \end{tabularx}

    \caption{Qualitative case study comparing the Ground Truth, the Comparison Method, the ablation variant without the $\Delta_{\text{prompt}}$, and the full model. Correctly identified pathologies are highlighted in \hlref{blue} and missed detections are marked with \hlfail{crosses}.}
    \label{fig:icon_comparison}
\end{figure}



\noindent\textbf{LLM-as-a-Judge: Multi-dimensional Pairwise Evaluation.}
 We use GPT-5 as a reference-guided judge for randomized pairwise comparisons in Figure~\ref{fig:llm} for multi-dimensional analysis. We evaluate descriptive quality using Granularity, Localization, and Quantification, which assess linguistic richness, anatomical specificity, and the use of measurable descriptors, respectively. We further assess clinical substance via Sensitivity and Utility, measuring coverage of reference findings and potential decision support. As shown in the radar chart, our method demonstrates a dominant advantage in descriptive quality, achieving superior win-rates across Granularity, Localization, and Quantification. This automated assessment provides a more detailed way of analyzing data, rather than replacing report generation metrics.

\noindent\textbf{Case Study}
Figure~\ref{fig:icon_comparison} shows qualitative comparisons. The comparison method, using a uniform input strategy, tends to generate template-like reports and miss subtle findings. In contrast, DiffVP uses reference-guided prompting to suppress redundant anatomy and emphasize informative evidence. The variant without $\Delta_{\text{prompt}}$ already captures linear subsegmental atelectasis missed by the baseline, while the full model with $\Delta_{\text{prompt}}$ further focuses on sparse deviations and recovers finer details such as the 4.6 mm measurement and steatosis.

\section{Conclusion}

In this paper, we propose DiffVP, a reference-guided framework that brings 3D CT report generation closer to the comparative reasoning used in clinical practice. DiffVP explicitly encodes scan-to-reference discrepancies and converts them into visual delta prompts via a hierarchical difference extractor and a difference-to-prompt generator. 
Injected as learnable prefixes to an LLM, these prompts provide structured visual guidance that suppresses redundant anatomical background and highlights informative evidence for report generation. Experiments show that DiffVP consistently surpasses state-of-the-art baselines in both language metrics and clinical efficacy. This work supports reference-guided prompting as an effective direction for medical report generation.

\section*{Acknowledgements}
This work is supported by the Natural Science Foundation of China under Grants 62271465 and 62502490; the National Key R\&D Program of China under Grant 2025YFC3408300; the Natural Science Foundation of Jiangsu Province under Grant BK20250496; the Suzhou Basic Research Program under Grant SYG202338; Jiangsu Funding Program for Excellent Postdoctoral Talent, and the China Postdoctoral Science Foundation under Grant 2024M763178.

\bibliographystyle{splncs04}
\bibliography{main}

\clearpage
\appendix

\section{Experiments Background}
\label{sec:appendix_background}

\subsection{Datasets}

\paragraph{RadGenome-ChestCT.}
Derived from the large-scale CT-RATE~\cite{hamamci2024foundation} dataset, which is specifically constructed for region-guided 3D CT interpretation.
RadGenome-ChestCT comprises a total of 25,692 CT-report pairs collected from 21,304 unique patients. We adhere to the official data partition, which designates 24,128 pairs for training and 1,564 pairs for evaluation.
This dataset includes overall CT data and uses SAT~\cite{zhao2023one} to segment the CT images into anatomical regions. Simultaneously, GPT-4o~\cite{achiam2023gpt} was used to localize and separate regions within the medical reports. The final output consists of image information and corresponding reports for ten common anatomical regions: abdomen, bones, breasts, esophagus, heart, lungs, trachea and bronchi, mediastinum, pleura, and thyroid. Furthermore,the dataset ensures standardized visual inputs with a consistent voxel spacing of $1 \times 1 \times 3$ mm.

\paragraph{CTRG-Chest-548K.}
Serving as a supplementary benchmark, this dataset contains 1,804 CT-report pairs.
To adapt the data for region-guided generation, we established a standardized preprocessing workflow.
We utilized SAT to extract visual masks for 10 anatomical regions, while leveraging Qwen-14B~\cite{yang2024qwen2} to parse and rewrite the raw reports into corresponding regional descriptions.
The dataset is randomly split into training and testing sets with a ratio of 8:2.

\subsection{Evaluation Metrics}

We evaluate generated reports using a combination of natural language generation (NLG) metrics and clinically oriented metrics.

For language quality, we report BLEU-$n$ ($n{=}1,2,3,4$)~\cite{papineni2002bleu},
METEOR~\cite{banerjee2005meteor} and ROUGE-L~\cite{lin2004rouge}.
BLEU-$n$ measures $n$-gram overlap between generated and reference reports,
ROUGE-L focuses on the longest common subsequence,
METEOR accounts for synonymy and paraphrasing, emphasizes consensus with multiple references by TF-IDF reweighting.
Together, these metrics assess lexical accuracy, fluency, and semantic similarity.

NLG metrics alone are insufficient to evaluate clinical correctness, as reports with opposite diagnostic meanings may still achieve similar scores.
To address this limitation, we additionally adopt \emph{clinical efficacy (CE)} metrics~\cite{hamamci2024foundation}.
Specifically, we use a RadBERT-based classifier~\cite{yan2022radbert} to extract 18 abnormality labels from both generated and ground-truth reports on the RadGenome-ChestCT dataset.
Precision, recall, and F1 scores are then computed over the predicted abnormality labels,
providing a clinically meaningful assessment of diagnostic consistency.

\section{Normal Reference Dataset}
\label{sec:normal_dataset}

To compare input images with normal reference CT images, we construct a dedicated normal reference sample dataset by selecting CT images whose reports explicitly indicate no abnormality.

\begin{figure}[t]
  \centering
  \small
  \setlength{\tabcolsep}{10pt}
  \renewcommand{\arraystretch}{1.25}
  \setlength{\arrayrulewidth}{0.6pt}

  \begin{tabularx}{\linewidth}
    {|>{\raggedright\arraybackslash}p{0.23\linewidth}
     |>{\raggedright\arraybackslash}X|}
    \hline
    \rowcolor{black}
    \multicolumn{2}{|c|}{\rule{0pt}{3.0ex}\color{white}\bfseries\large Normal Reference Report} \\  
    \hline

    \rowcolor{tblgray}
    \textbf{Dataset} & \textbf{Reference Content} \\
    \hline

    \rowcolor{tblgray}
    RadGenome &
    Thoracic CT examination within normal limits.\par
    Findings within normal limits.\par
    No obvious pathology was observed in thorax CT examination.\par
    Findings within normal limits
    \\
    \hline

    \rowcolor{tblgray}
    CTRG-548K &
    Thorax is symmetrical. lung window shows clear lung markings. natural walking. good lung field transparency. no obvious consolidation shadow. and bilateral pulmonary hili are not large. The shape of heart shadow and heart big vessels is normal. and no obvious mass or enlarged lymph node is found in mediastinum. No pleural effusion or pleural thickening.\par
    Thorax is symmetrical. lung window shows clear lung markings. natural walking. good lung field transparency. no obvious consolidation shadow. and bilateral pulmonary hili are not large. The shape of heart shadow and heart big vessels is normal. and no obvious mass or enlarged lymph node is found in mediastinum. No pleural effusion or pleural thickening.
    \\
    \hline
  \end{tabularx}

  \caption{Examples of normal references from the constructed pool. These reports describe standard physiological states without pathological findings.}
  \label{fig:normal_examples}
\end{figure}

This selection process yields 1,787 training samples and 124 evaluation samples from the RadGenome-ChestCT dataset, and 105 training samples and 43 test samples from the CTRG-Chest-548K dataset.

During the training and inference phases, we employ a random pairing strategy, matching each target input image with a reference sample randomly drawn from the corresponding portion of this normal reference dataset. Note that there is no mixing or leakage of normal reference images used in the training and testing processes; the normal reference samples drawn from each dataset, belonging to both the training and test phases, were used only for the corresponding dataset's training or testing phase.

Since the normal reference samples were collected based on reports, we provide examples of normal reports from the two datasets below to give a more intuitive understanding of normal samples in figure~\ref{fig:normal_examples}.

\section{Stability and Significance Analysis}
\label{sec:stability_significance}

\begin{table}[t]
\centering
\scriptsize
\setlength{\tabcolsep}{3.2pt}
\renewcommand{\arraystretch}{1.05}
\caption{
Quantitative results and stability analysis on the test set.
We report results across five random seeds for both Baseline and Ours, together with mean$\pm$std and two-sided $t$-test $p$-values (Ours vs.\ Baseline).
Metric scores are reported in percentage (\%).
}
\label{tab:significance_5seeds_updated}

\resizebox{\linewidth}{!}{%
\begin{tabular}{lccccccc}
\toprule
\textbf{Method} & \textbf{B1} & \textbf{B4} & \textbf{MTR} & \textbf{RG-L} & \textbf{F1} & \textbf{Recall} & \textbf{Precision} \\
\midrule
Baseline (Seed 21) & 47.25 & 24.87 & 44.07 & 36.65 & 0.253 & 0.180 & 0.423 \\
Baseline (Seed 22) & 47.43 & 25.02 & 44.06 & 36.67 & 0.250 & 0.178 & 0.417 \\
Baseline (Seed 23) & 47.63 & 25.27 & 44.38 & 36.98 & 0.248 & 0.178 & 0.412 \\
Baseline (Seed 24) & 47.41 & 25.09 & 44.18 & 37.62 & 0.260 & 0.186 & 0.430 \\
Baseline (Seed 25) & 47.30 & 24.84 & 43.99 & 36.53 & 0.259 & 0.186 & 0.427 \\
\midrule
Ours (Seed 21) & 58.01 & 35.82 & 50.76 & 39.48 & 0.398 & 0.440 & 0.362 \\
Ours (Seed 22) & 57.75 & 35.54 & 50.59 & 39.27 & 0.397 & 0.441 & 0.361 \\
Ours (Seed 23) & 57.73 & 35.54 & 50.59 & 39.28 & 0.397 & 0.441 & 0.361 \\
Ours (Seed 24) & 57.75 & 35.52 & 50.58 & 39.25 & 0.397 & 0.441 & 0.361 \\
Ours (Seed 25) & 58.16 & 34.28 & 47.40 & 32.32 & 0.406 & 0.450 & 0.370 \\
\midrule
\textbf{Baseline (mean$\pm$std)} &
47.40{\scriptsize$\pm$0.15} &
25.02{\scriptsize$\pm$0.17} &
44.14{\scriptsize$\pm$0.15} &
36.89{\scriptsize$\pm$0.39} &
0.254{\scriptsize$\pm$0.005} &
0.182{\scriptsize$\pm$0.004} &
0.422{\scriptsize$\pm$0.007} \\
\textbf{Ours (mean$\pm$std)} &
57.88{\scriptsize$\pm$0.19} &
35.34{\scriptsize$\pm$0.61} &
49.98{\scriptsize$\pm$1.45} &
37.92{\scriptsize$\pm$3.13} &
0.399{\scriptsize$\pm$0.003} &
0.443{\scriptsize$\pm$0.004} &
0.363{\scriptsize$\pm$0.004} \\
\textbf{$p$-value} &
8.77e-13 & 6.44e-07 & 7.65e-04 & 5.05e-01 & 1.62e-10 & 1.13e-13 & 3.36e-06 \\
\bottomrule
\end{tabular}%
}
\end{table}

\paragraph{Stability under Random Test-Time Pairing.}
Our framework pairs each abnormal input with a randomly sampled normal reference at inference time, reflecting realistic clinical conditions where perfectly aligned priors are unavailable. To evaluate robustness to this stochastic pairing, we run both the Baseline and our method five times (Seeds 21--25), each with an independent random pairing.

\paragraph{Robustness to Stochastic Pairing.}
Table~\ref{tab:significance_5seeds_updated} summarizes the cross-seed mean$\pm$std. Despite random reference selection, our method remains stable across runs, with small variance on both language and clinical metrics (eg\ std$=0.19$ for BLEU-1 and std$=0.003$ for CE-F1). This indicates that the difference-aware mechanism extracts consistent semantic discrepancies rather than relying on specific reference samples or ``lucky'' pairings.

\paragraph{Statistical Significance.}
We further test whether the improvements are statistically reliable using two-sided $t$-tests over the five seeds. As reported in Table~\ref{tab:significance_5seeds_updated}, the gains are significant for BLEU-1, BLEU-4, METEOR, and clinical efficacy metrics (F1/Recall/Precision), with $p<10^{-3}$. In contrast, ROUGE-L does not show a significant difference ($p=5.05\mathrm{e}{-01}$), consistent with the observation that longest-common-subsequence structure can be more sensitive to stochastic generation. Overall, these results support that our improvements are robust and reproducible under random test-time pairing.

\paragraph{Why ROUGE-L can be non-significant.}
ROUGE-L is computed from the longest common subsequence (LCS) and is therefore sensitive to sentence order and discourse structure: when the same clinical content is organized in a different order, the LCS length can decrease even if the semantics remain consistent. We inspected predictions against ground-truth reports and found that the weaker and non-significant ROUGE-L results mainly stem from structural ordering differences rather than missing clinical content. 

For example, in case \texttt{valid\_1289}, the ground truth follows the order
\emph{abdomen $\rightarrow$ bone $\rightarrow$ esophagus $\rightarrow$ heart and mediastinum $\rightarrow$ lung $\rightarrow$ lymph nodes $\rightarrow$ airway},
whereas our report follows
\emph{abdomen $\rightarrow$ bone $\rightarrow$ esophagus $\rightarrow$ mediastinal vessels $\rightarrow$ lung $\rightarrow$ mediastinum and lymph nodes $\rightarrow$ airway}.

Although the anatomical coverage and abnormal findings largely overlap, differences in section placement and topic order shorten the LCS alignment, lowering ROUGE-L. Hence, the non-significant ROUGE-L difference reflects structural sensitivity rather than semantic deficiency, which is consistent with the stable clinical efficacy metrics. We acknowledge that structure alignment can be further improved and will explore structure-aware strategies in future work.

\section{Diagnostic Anchor $E$: Definition, Training, and Ablations}
\label{sec:supp_E}

\paragraph{Definition and supervision source.}
We introduce an auxiliary diagnostic anchor $E$ as an 18-class multi-label disease vector aligned with the official structured annotations of RadGenome-ChestCT.

Each CT volume corresponds to an 18-class label over the following categories:
\emph{Medical material, Arterial wall calcification, Cardiomegaly, Pericardial effusion, Coronary artery wall calcification, Hiatal hernia, Lymphadenopathy, Emphysema, Atelectasis, Lung nodule, Lung opacity, Pulmonary fibrotic sequela, Pleural effusion, Mosaic attenuation pattern, Peribronchial thickening, Consolidation, Bronchiectasis, Interlobular septal thickening}.
These labels are dataset-provided structured annotations rather than extracted from free-text reports.
The $E$ classifier is trained only on the training split, hence it does not introduce information leakage.

\paragraph{Training of the $E$ predictor.}
$E$ is predicted by an independently trained 3D multi-label classifier with a 3D ResNet-18 backbone.
The classifier produces voxel-level logits, which are aggregated into volume-level probabilities using Noisy-OR pooling.
We optimize the classifier using a standard multi-label binary cross-entropy (BCE) loss with the 18-class labels.

\paragraph{How $E$ is used for report generation.}
During report generation, the $E$ classifier is frozen.
Predicted disease probabilities are converted into short diagnostic prompts and prepended to the LLM input as a lightweight semantic anchor.
We also tested end-to-end joint training of the classifier and the generator, but observed unstable optimization and reduced report quality; therefore, we adopt a two-stage design.
Using predicted disease labels as semantic anchors is a standard paradigm in medical report generation and vision--language modeling, and is not the core novelty of this work.

\paragraph{Ablation without $E$.}
To evaluate the standalone effect of difference-aware modeling, we remove $E$ and report results in Table~\ref{tab:ablation_noE}.
Even without $E$, discrepancy modeling still yields consistent improvements over the baseline.
In particular, introducing $\Delta_{\text{global}}$ improves overall language generation quality across NLG metrics, and further adding $\Delta_{\text{local}}$ enhances abnormality sensitivity, increasing CE-F1 from 0.188 to 0.238 and improving Recall.
These results indicate that multi-level discrepancy modeling has an independent effect and contributes directly to clinical effectiveness, rather than relying on the diagnostic anchor.

\begin{table}[t]
\centering
\small
\setlength{\tabcolsep}{4pt}
\renewcommand{\arraystretch}{1.05}
\caption{Ablation study without the diagnostic anchor $E$ on RadGenome-ChestCT.}
\label{tab:ablation_noE}
\begin{tabular}{lccccccc}
\toprule
Setting & BL-1 & BL-4 & MTR & RG-L & F1 & Recall & Precision \\
\midrule
Baseline (w/o $\Delta$, w/o $E$) & 55.29 & 33.17 & 44.11 & 31.94 & 0.159 & 0.109 & 0.295 \\
$+\Delta_{\text{global}}$ (w/o $E$) & 56.83 & 37.24 & 50.09 & 42.50 & 0.188 & 0.124 & 0.383 \\
$+\Delta_{\text{global}}+\Delta_{\text{local}}$ (w/o $E$) & 57.23 & 33.99 & 46.67 & 32.83 & 0.238 & 0.175 & 0.374 \\
$+\Delta_{\text{global}}+\Delta_{\text{local}}+E$ (Ours) & 58.16 & 34.28 & 47.40 & 32.32 & 0.421 & 0.496 & 0.366 \\
\bottomrule
\end{tabular}
\end{table}

\section{LLM-as-Judge Evaluation Protocol}
\label{sec:llm_judge}

We use GPT-5 (temperature = 0) as a reference-guided judge for randomized pairwise comparisons across five clinically motivated dimensions: finding coverage, clinical actionability, detail richness, anatomical specificity, and quantification detail. The evaluation follows a forced-choice design to compute win-rates. We include the complete judge prompt below to ensure full reproducibility and transparency.

\section{Sensitivity to Normal Reference Pool}
\label{sec:supp_ref_sensitivity}

This section evaluates the sensitivity of our method to (i) the size of the normal reference pool, (ii) domain shift in the normal references, and (iii) imperfect normal pools with abnormal contamination, following the reviewer request to vary pool size and pool quality and report performance.

\begin{center}
  \setlength{\tabcolsep}{8pt}
  \renewcommand{\arraystretch}{1.10}
  \setlength{\arrayrulewidth}{0.6pt}
  \small

  \begin{tabularx}{\linewidth}{|X|}
    \hline
    \rowcolor{black}
    \multicolumn{1}{|c|}{\rule{0pt}{3.0ex}\color{white}\bfseries\large Judge Prompt (GPT-5)} \\
    \hline

    \rowcolor{tblgray}
    \begin{minipage}[t]{0.98\linewidth}
    \vspace{4pt}
    \scriptsize
    \raggedright
    \sloppy

You are a Senior Radiologist Consultant.

You must compare Report A vs Report B against the Ground Truth (GT) on FIVE dimensions.
IMPORTANT: You are NOT allowed to output "Tie" in any dimension. You must choose either "Report A" or "Report B" for every winner field.
This forced-choice design is required to compute consistent win-rate statistics.

Focus strictly on what is written in the reports. Do not assume extra findings beyond GT.
When judging "detail\_richness", richness should NOT reward fabricated or unsupported facts. Prefer clinically plausible elaboration only if it remains consistent with the GT. Penalize clearly invented or contradictory details.

\medskip
\textbf{Dimension 1: finding\_coverage}\\
\textbf{Question:} Which report captured more GT POSITIVE findings (missed fewer true abnormalities)?\\
\textbf{How to judge:}\\
\textbullet\ Identify GT positive abnormalities and check which report mentions them correctly.\\
\textbullet\ Missing a major GT abnormality is heavily penalized.\\
\textbf{Winner:} the report that misses fewer true positives.

\medskip
\textbf{Dimension 2: clinical\_actionability}\\
\textbf{Question:} Which report is more clinically actionable while staying faithful to GT?\\
\textbf{How to judge:}\\
\textbullet\ Clear key impression / prioritization of important findings.\\
\textbullet\ Less irrelevant clutter; avoids burying key abnormalities.\\
\textbullet\ No major contradictions with GT.\\
\textbf{Winner:} the report that is more useful for clinical decision-making.

\medskip
\textbf{Dimension 3: detail\_richness}\\
\textbf{Question:} Which report provides richer clinically-relevant detail while remaining faithful to GT?\\
\textbf{How to judge:}\\
\textbullet\ More descriptive radiology language, qualifiers, and contextual details.\\
\textbullet\ Avoid empty generic statements.\\
\textbullet\ Prefer plausible and GT-consistent details.\\
\textbullet\ Penalize unsupported measurements, invented facts, or contradictions.\\
\textbf{Winner:} the report with richer and informative yet GT-consistent descriptions.

\medskip
\textbf{Dimension 4: anatomical\_specificity}\\
\textbf{Question:} Which report uses more specific anatomical localization?\\
\textbf{How to judge:}\\
\textbullet\ Laterality (left/right), lobes/segments, sub-regions, organ parts.\\
\textbullet\ Concrete localization beats vague phrases.\\
\textbf{Winner:} the report with finer anatomical grounding.

\medskip
\textbf{Dimension 5: quantification\_detail}\\
\textbf{Question:} Which report includes more quantification and measurable descriptors?\\
\textbf{How to judge:}\\
\textbullet\ Measurements (mm/cm), counts (multiple/few), severity (mild/moderate/severe), intervals.\\
\textbf{Winner:} the report with more quantification.

\medskip
\textbf{Output format (MUST be a strict JSON object, no markdown, no extra text):}\\
{\ttfamily
\{
  "finding\_coverage\_winner": "Report A" or "Report B",
  "clinical\_actionability\_winner": "Report A" or "Report B",
  "detail\_richness\_winner": "Report A" or "Report B",
  "anatomical\_specificity\_winner": "Report A" or "Report B",
  "quantification\_detail\_winner": "Report A" or "Report B",

  "evidence": \{
    "GT\_key\_findings": ["<=5 short phrases summarizing the GT positives, if any; if GT is fully normal, write ['normal']"],
    "ReportA\_supporting\_spans": ["2-4 short verbatim spans copied from Report A that support your decisions"],
    "ReportB\_supporting\_spans": ["2-4 short verbatim spans copied from Report B that support your decisions"]
  \},

  "reason": "1-3 sentences summarizing why the winners were chosen across the five dimensions."
\}
}

    \vspace{4pt}
    \end{minipage}
    \\
    \hline
  \end{tabularx}

    \captionof{figure}{Complete prompt used for LLM-as-judge evaluation (GPT-5, temperature$=0$).}
  \label{fig:judge_prompt}
\end{center}

\subsection{Effect of Normal Reference Pool Size}
\label{sec:supp_pool_size}

\begin{table}[t]
\centering
\small
\setlength{\tabcolsep}{4pt}
\renewcommand{\arraystretch}{1.05}
\caption{Effect of normal reference pool size on RadGenome-ChestCT.}
\label{tab:pool_size}
\begin{tabular}{c c c c c c c c}
\toprule
\textbf{Pool Size} & \textbf{B1} & \textbf{B4} & \textbf{M} & \textbf{RL} & \textbf{F1} & \textbf{Recall} & \textbf{Prec} \\
\midrule
1   & 58.20 & 35.82 & 50.71 & 39.38 & 0.403 & 0.451 & 0.365 \\
10  & 58.33 & 36.07 & 50.83 & 39.50 & 0.406 & 0.451 & 0.370 \\
40  & 58.40 & 36.31 & 50.92 & 39.74 & 0.414 & 0.458 & 0.379 \\
60  & 58.25 & 35.89 & 50.66 & 39.47 & 0.403 & 0.450 & 0.365 \\
124 & 58.08 & 35.73 & 50.55 & 39.38 & 0.398 & 0.445 & 0.360 \\
\bottomrule
\end{tabular}
\end{table}

We conduct additional experiments on RadGenome-ChestCT using different normal reference pool sizes: 1, 10, 40, 60, and 124 images (124 corresponds to the full reference pool in the Rad dataset). As shown in Table~\ref{tab:pool_size}, both NLG metrics and clinical efficiency metrics remain stable across different pool sizes. The overall variation is small.

This behavior is closely related to our random pairing design. During both training and testing, abnormal scans are randomly paired with normal references rather than matched to fixed samples. Therefore, the model learns a distributional representation of abnormal--normal contrast patterns instead of memorizing specific reference images or relying on a large reference pool. Even with a small pool, the stochastic sampling mechanism enables the model to consistently extract stable difference signals.

Overall, the results indicate that the proposed method is robust to the size of the normal reference pool and does not depend on a large curated reference library.

\subsection{Effect of Domain Shift}
\label{sec:supp_domain_shift}

To evaluate the impact of domain shift, we construct a cross-domain normal pool. Abnormal samples remain from the RadGenome-ChestCT test set, while normal reference images are replaced with normal cases from the CTRG dataset. CTRG and RadGenome differ systematically in scanning protocols, preprocessing pipelines, pixel statistics, reconstruction orientation, contrast usage, device types, hospital sources, and acquisition periods. Therefore, abnormal--normal pairing under this setting constitutes a realistic cross-domain scenario.

\begin{table}[t]
\centering
\small
\setlength{\tabcolsep}{4pt}
\renewcommand{\arraystretch}{1.05}
\caption{In-domain vs.\ domain-shift comparison on RadGenome-ChestCT.}
\label{tab:domain_shift_main}
\begin{tabular}{l c c c c c c c}
\toprule
\textbf{Setting} & \textbf{B1} & \textbf{B4} & \textbf{M} & \textbf{RL} & \textbf{F1} & \textbf{Recall} & \textbf{Prec} \\
\midrule
Baseline (No $\Delta$)    & 55.29 & 33.17 & 44.11 & 31.94 & 0.159 & 0.109 & 0.295 \\
Ours (In-domain $\Delta$) & 58.16 & 34.28 & 47.40 & 32.32 & 0.421 & 0.496 & 0.366 \\
Ours (100\% CTRG $\Delta$) & 56.53 & 34.95 & 48.03 & 36.50 & 0.128 & 0.082 & 0.289 \\
\bottomrule
\end{tabular}
\end{table}

Under domain shift, B1 decreases from 58.16 (in-domain $\Delta$) to 56.53, but remains clearly higher than the no-$\Delta$ baseline (55.29). This indicates that domain shift weakens the contrast signal to some extent, yet the $\Delta$ mechanism still provides consistent gains.

We further evaluate 10\%, 50\%, and 100\% CTRG mixing ratios. NLG metrics remain relatively stable without monotonic degradation as the mixing ratio increases, suggesting reasonable robustness to cross-domain normal references.

\begin{table}[t]
\centering
\small
\setlength{\tabcolsep}{4pt}
\renewcommand{\arraystretch}{1.05}
\caption{Domain shift with different CTRG mixing ratios.}
\label{tab:domain_shift_mix}
\begin{tabular}{c c c c c c c c}
\toprule
\textbf{CTRG Mix Ratio} & \textbf{B1} & \textbf{B4} & \textbf{M} & \textbf{RL} & \textbf{F1} & \textbf{Recall} & \textbf{Prec} \\
\midrule
10\%  & 56.73 & 35.05 & 48.17 & 36.66 & 0.130 & 0.083 & 0.299 \\
50\%  & 56.91 & 35.45 & 48.47 & 36.82 & 0.128 & 0.082 & 0.292 \\
100\% & 56.53 & 34.95 & 48.03 & 36.50 & 0.128 & 0.082 & 0.289 \\
\bottomrule
\end{tabular}
\end{table}

Clinical efficiency (CE) metrics are more sensitive to domain shift. For example, F1 drops from 0.421 (in-domain) to approximately 0.128 under full cross-domain setting. This is expected, as CE directly depends on the precision of abnormal--normal contrast, which is affected by distribution mismatch.

Importantly, in practical deployment, obtaining a small number of in-domain normal samples is typically low-cost. Combined with our earlier findings that performance is stable even with small normal pools, domain shift is unlikely to be a major obstacle in real-world application.

\subsection{Imperfect Normal Pool}
\label{sec:supp_imperfect_pool}

To evaluate the impact of an imperfect normal pool, we contaminate the normal reference set by mixing abnormal samples at ratios of 10 percent and 50 percent. Evaluation is conducted on the same test set.

Results show that NLG metrics such as BLEU, ROUGE, and METEOR remain relatively stable, while CE metrics decline. For example, F1 decreases from 0.421 under a clean normal pool to 0.166 at 10 percent contamination and 0.128 at 50 percent contamination.

This indicates that when the normal pool is no longer clean, the abnormal--normal contrast signal weakens, primarily affecting abnormality-related metrics. However, overall text generation quality remains stable. This experiment directly quantifies performance changes under imperfect normal conditions.

\begin{table}[t]
\centering
\small
\setlength{\tabcolsep}{5pt}
\renewcommand{\arraystretch}{1.05}
\caption{Imperfect normal pool with mixed abnormal samples.}
\label{tab:imperfect_pool}
\begin{tabular}{l c c c c c c c}
\toprule
\textbf{Abnormal Ratio} & \textbf{B1} & \textbf{B4} & \textbf{M} & \textbf{RL} & \textbf{F1} & \textbf{Recall} & \textbf{Prec} \\
\midrule
10\%      & 57.03 & 36.51 & 48.41 & 36.34 & 0.166 & 0.118 & 0.282 \\
50\%      & 56.91 & 35.45 & 48.47 & 36.82 & 0.128 & 0.082 & 0.292 \\
0\% (ours) & 58.16 & 34.28 & 47.40 & 32.32 & 0.421 & 0.496 & 0.366 \\
\bottomrule
\end{tabular}
\end{table}

\subsection{Summary}
\label{sec:supp_ref_summary}

We systematically evaluate three scenarios: small normal pools, cross-domain normal pools, and imperfect normal pools with mixed abnormal samples. Results show that performance remains stable under small pool sizes. Under domain shift, performance decreases moderately but remains above the no-$\Delta$ baseline. When the normal pool is contaminated, abnormality-related metrics decline, while generation quality remains largely stable.

These findings indicate that the method is robust to variations in normal pool size and distribution, although abnormal contrast capability weakens as reference quality degrades.

In practical deployment, obtaining a small number of in-domain and relatively clean normal samples is typically low cost. Only a limited reference set is needed to support the contrast-based mechanism. Therefore, reliance on a curated normal pool does not pose a substantial limitation for real-world application.

%
%
\end{document}